\documentclass[10pt,twocolumn,letterpaper]{article}

\usepackage[pagenumbers]{cvpr}

\usepackage{graphicx}
\usepackage{amsmath}
\usepackage{amssymb}
\usepackage{booktabs}
\usepackage{multirow}
\usepackage{multicol}
\usepackage{subcaption}
\usepackage{amsfonts}
\usepackage{comment}
\usepackage{color}
\usepackage[accsupp]{axessibility}
\usepackage{pifont}
\usepackage[pagebackref,breaklinks,colorlinks]{hyperref}

\usepackage[capitalize]{cleveref}
\crefname{section}{Sec.}{Secs.}
\Crefname{section}{Section}{Sections}
\Crefname{table}{Table}{Tables}
\crefname{table}{Tab.}{Tabs.}

\def\confName{CVPR}
\def\confYear{2023}

\begin{document}

\title{DivClust: Controlling Diversity in Deep Clustering}

\author{Ioannis Maniadis Metaxas\thanks{Corresponding author}, Georgios Tzimiropoulos, Ioannis Patras\\
Queen Mary University of London\\
Mile End road, E1 4NS London, UK\\
{\tt\small \{i.maniadismetaxas, g.tzimiropoulos, i.patras\}@qmul.ac.uk}
}

\maketitle

\begin{abstract}
Clustering has been a major research topic in the field of machine learning, one to which Deep Learning has recently been applied with significant success. However, an aspect of clustering that is not addressed by existing deep clustering methods, is that of efficiently producing multiple, diverse partitionings for a given dataset. This is particularly important, as a diverse set of base clusterings are necessary for consensus clustering, which has been found to produce better and more robust results than relying on a single clustering. To address this gap, we propose DivClust, a diversity controlling loss that can be incorporated into existing deep clustering frameworks to produce multiple clusterings with the desired degree of diversity. We conduct experiments with multiple datasets and deep clustering frameworks and show that: a) our method effectively controls diversity across frameworks and datasets with very small additional computational cost, b) the sets of clusterings learned by DivClust include solutions that significantly outperform single-clustering baselines, and c) using an off-the-shelf consensus clustering algorithm, DivClust produces consensus clustering solutions that consistently outperform single-clustering baselines, effectively improving the performance of the base deep clustering framework. Code is available at \url{https://github.com/ManiadisG/DivClust}.
\end{abstract}

\setlength{\tabcolsep}{6pt}

\section{Introduction}\label{sec:intro}

\begin{figure*}[t]
\begin{center}
\includegraphics[width=1.\linewidth]{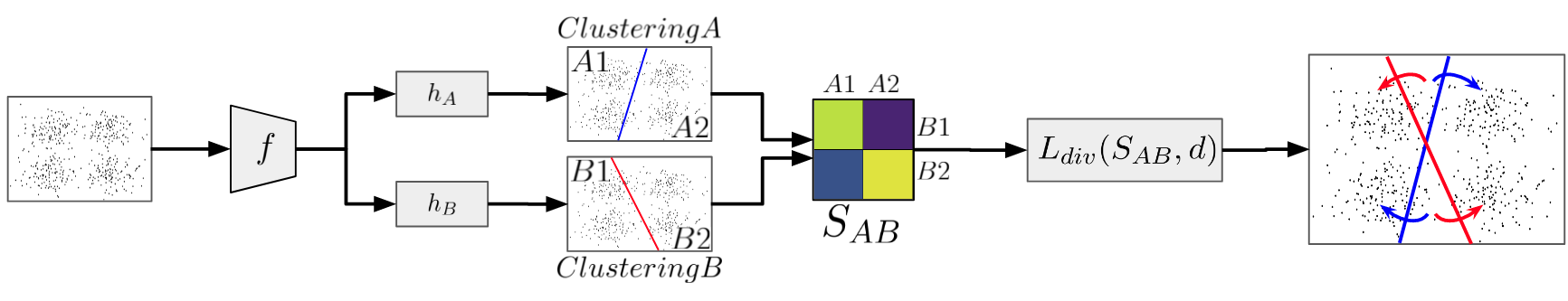}
\end{center}
\caption{Overview of DivClust. Assuming clusterings $A$ and $B$, the proposed diversity loss $L_{div}$ calculates their similarity matrix $S_{AB}$ and restricts the similarity between cluster pairs to be lower than a similarity upper bound $d$. In the figure, this is represented by the model adjusting the cluster boundaries to produce more diverse clusterings. Best seen in color.}
\label{fig:overview}
\end{figure*}

The exponentially increasing volume of visual data, along with advances in computing power and the development of powerful Deep Neural Network architectures, have revived the interest in unsupervised learning with visual data. Deep clustering in particular has been an area where significant progress has been made in the recent years. Existing works focus on producing a single clustering, which is evaluated in terms of how well that clustering matches the ground truth labels of the dataset in question. However, consensus, or ensemble, clustering remains under-studied in the context of deep clustering, despite the fact that it has been found to consistently improve performance over single clustering outcomes~\cite{ensemble_survey,zhou2021self,ghaemi2009survey,liu2019consensus}.

Consensus clustering consists of two stages, specifically generating a set of base clusterings, and then applying a consensus algorithm to aggregate them. Identifying what properties ensembles should have in order to produce better outcomes in each setting has been an open problem~\cite{golalipour2021clustering}. However, research has found that inter-clustering diversity within the ensemble is an important, desirable factor~\cite{pividori2016diversity,hamidi2022impact,gullo2009diversity,fern2003random,iam2011link}, along with individual clustering quality, and that diversity should be moderated~\cite{moderate,moderate2,pividori2016diversity}. Furthermore, several works suggest that controlling diversity in ensembles is important toward studying its impact and determining its optimal level in each setting~\cite{moderate,pividori2016diversity}. 

The typical way to produce diverse clusterings is to promote diversity by clustering the data multiple times with different initializations/hyperparameters or subsets of the data~\cite{ghaemi2009survey,ensemble_survey}. This approach, however, does not guarantee or control the degree of diversity, and is computationally costly, particularly in the context of deep clustering, where it would require the training of multiple models. Some methods have been proposed that find diverse clusterings by including diversity-related objectives to the clustering process, but those methods have only been applied to clustering precomputed features and cannot be trivially incorporated into Deep Learning frameworks. Other methods tackle diverse clustering by creating and clustering diverse feature subspaces, including some that apply this approach in the context of deep clustering~\cite{DiMVMC,enrc}. Those methods, however, do not control inter-clustering diversity. Rather, they influence it indirectly through the properties of the subspaces they create. Furthermore, typically, existing methods have been focusing on producing orthogonal clusterings or identifying clusterings based on independent attributes of relatively simple visual data (e.g. color/shape). Consequently, they are oriented toward \textit{maximizing} inter-clustering diversity, which is not appropriate for consensus clustering~\cite{moderate,moderate2,pividori2016diversity}.

To tackle this gap, namely generating multiple clusterings with deep clustering frameworks efficiently and with the desired degree of diversity, we propose DivClust. Our method can be straightforwardly incorporated into existing deep clustering frameworks to learn multiple clusterings whose diversity is \textit{explicitly controlled}. Specifically, the proposed method uses a single backbone for feature extraction, followed by multiple projection heads, each producing cluster assignments for a corresponding clustering. Given a user defined diversity target, in this work expressed in terms of the average NMI between clusterings, DivClust restricts inter-clustering similarity to be below an appropriate, dynamically estimated threshold. This is achieved with a novel loss component, which estimates inter-clustering similarity based on soft cluster assignments produced by the model, and penalizes values exceeding the threshold. Importantly, DivClust introduces minimal computational cost and requires no hyperparameter tuning with respect to the base deep clustering framework, which makes its use simple and computationally efficient.

Experiments on four datasets (CIFAR10, CIFAR100, Imagenet-10, Imagenet-Dogs) with three recent deep clustering methods (IIC~\cite{IIC}, PICA~\cite{PICA}, CC~\cite{li2021contrastive}) show that DivClust can effectively control inter-clustering diversity without reducing the quality of the clusterings. Furthermore, we demonstrate that, with the use of an off-the-shelf consensus clustering algorithm, the diverse base clusterings learned by DivClust produce consensus clustering solutions that outperform the base frameworks, effectively improving them with minimal computational cost. Notably, despite the sensitivity of consensus clustering to the properties of the ensemble, our method is robust across various diversity levels, outperforming baselines in most settings, often by large margins. Our work then provides a straightforward way for improving the performance of deep clustering frameworks, as well as a new tool for studying the impact of diversity in deep clustering ensembles~\cite{pividori2016diversity}.

In summary, DivClust: a) can be incorporated in existing deep clustering frameworks in a plug-and-play way with very small computational cost, b) can explicitly and effectively control inter-clustering diversity to satisfy user-defined targets, and c) learns clusterings that can improve the performance of deep clustering frameworks via consensus clustering.

\section{Related Works}\label{sec:related_works}

\subsection{Deep Clustering}
The term deep clustering refers to methods that cluster data while learning their features. They are generally divided into two categories, namely those that alternate training between clustering and feature learning and those that train both simultaneously.

\textbf{Alternate learning}: Methods following this approach generally utilize a two-step training regime repeated in regular intervals (e.g. per-epoch or per-step). First, sample pseudo-labels are produced based on representations extracted by the model (e.g. by feature clustering). Second, those pseudo-labels are utilized to improve the learned representations, typically by training the feature extraction model as a classifier. Those methods include DEC~\cite{DEC}, DAC~\cite{DAC}, DCCM~\cite{DCCM}, DDC~\cite{DDC}, JULE~\cite{JULE}, SCAN~\cite{SCAN}, ProPos\cite{huang2022learning} and SPICE~\cite{niu2022spice}, as well as DSC-N~\cite{DSC-N}, IDFD~\cite{tao2021clustering} and MIX'EM~\cite{MIXEM}, which propose ways to train models whose representations produce better outcomes when clustered. Other works in this area are DeepCluster~\cite{deepcluster}, SeLa~\cite{sela}, PCL~\cite{li2020prototypical} and HCSC~\cite{guo2022hcsc}, though their primary focus is on representation learning.

\textbf{Simultaneous learning}: These methods jointly learn features and cluster assignments. They include ADC~\cite{ADC}, IIC~\cite{IIC} and PICA~\cite{PICA}, which train clustering models end-to-end with loss functions that enforce desired properties on the clusters assignments, ConCURL~\cite{regatti2021consensus,deshmukh2022representation}, which builds on BYOL~\cite{grill2020bootstrap} with a loss maximizing the agreement of clusterings from transformed embeddings, DCCS~\cite{DCCS}, which leverages an adversarial component in the clustering process, and GatCluster~\cite{GatCluster}, which proposes an attention mechanism combined with four self-learning tasks. Finally, methods such as SCL~\cite{scl}, CC~\cite{li2021contrastive}, GCC~\cite{GCC}, TCC~\cite{shen2021you} and MiCE~\cite{MiCE} leverage contrastive learning.

Although some deep clustering methods~\cite{sela,IIC,regatti2021consensus} use multiple clusterings, most do not explore the prevalence and impact of inter-clustering diversity, and none proposes ways to control it. Our work is, to the best of our knowledge, the first that addresses both issues.

\subsection{Diverse Clustering}
The most straightforward way of producing multiple, diverse clusterings is clustering the data multiple times. Typical methods to increase diversity include varying the clustering algorithm or its hyperparameters, using different initializations, and clustering a subset of the samples or features~\cite{ensemble_survey}. This approach, however, is a) computationally costly, in that it requires clustering the data multiple times, b) unreliable, as some ways to increase diversity might decrease the quality of clusters (e.g. using a subset of the data), and c) ineffective, as there is no guarantee that the desired degree of diversity will be achieved.

To tackle this, several methods have been proposed to create multiple, diverse clusterings~\cite{hu2018subspace}. We identify two main approaches of promoting inter-clustering diversity: a) explicitly, by optimizing for appropriate objectives, and b) implicitly, by optimizing for decorrelated/orthogonal feature subspaces, which, when clustered, lead to diverse clusterings. Methods in the first category include COALA~\cite{COALA}, Meta Clustering~\cite{METAC}, \mbox{Dec-kmeans}~\cite{Dec-kmeans}, MNMF~\cite{MNMF}, MSC~\cite{MSC}, ADFT~\cite{ADFT} and MultiCC~\cite{MULTICC}. Subspace clustering methods include MISC~\cite{MISC}, ISAAC~\cite{ISAAC}, NR-kmeans~\cite{NRkmeans}, RAOSC~\cite{raosc} and ENRC~\cite{enrc}. Distinctly, diverse clustering has also been explored in the context of multi-view data by OSC~\cite{OSC}, MVMC~\cite{MVMC}, DMSMF~\cite{DMSMF}, DMClusts~\cite{DMClusts}, DiMSC~\cite{DiMSC}, and DiMVMC~\cite{DiMVMC}.

To the best of our knowledge, except for DiMVMC and ENRC, \textit{none} of the existing methods are compatible with Deep Learning, they require a learned feature space on which to be applied, and most have quadratic complexity relative to the number of samples. This restricts their use on real-life high dimensional data, where deep clustering produces better outcomes~\cite{MiCE,SCAN}. Regarding DiMVMC and ENRC, they depend on autoencoder-based architectures and adapting them to more recent deep clustering frameworks, which perform significantly better, is not trivial. More importantly, they utilize subspace clustering, inheriting its limitations regarding controlling diversity. Specifically:  a) no method has been proposed to infer \textit{how} different the subspaces must be in order to lead to a \textit{specific} degree of inter-clustering diversity, and b) subspace clustering methods inherit the randomness of the clustering algorithm applied to the subspaces (K-means for DiMVMC and ENRC), which further limits their control over the outcomes.

\subsection{Consensus Clustering}

The performance of clustering algorithms varies depending on the data and their properties, the algorithm itself, and its hyperparameters. This makes finding reliable clustering solutions particularly difficult. Consensus, or ensemble, clustering has emerged as a solution to this problem, specifically by combining the results of multiple, different clusterings, rather than relying on a single solution. This has been found to produce better and more robust outcomes than single-clustering approaches~\cite{ensemble_survey,golalipour2021clustering,ghaemi2009survey,liu2019consensus}. The process of consensus clustering happens in two stages: a) multiple, diverse base clusterings are generated and b) those clusterings are aggregated using a consensus algorithm. 

\textbf{Generating diverse clusterings:} The properties of the set of clusterings used by the consensus algorithm is a key factor for obtaining good performance. Multiple works~\cite{pividori2016diversity,hamidi2022impact,kuncheva2004using} have found that both the quality of individual base clusterings and their diversity is critical, and that, indeed, clustering ensembles with a moderate degree of diversity lead to better outcomes~\cite{moderate,moderate2,gullo2009diversity}. Typical methods for ensemble generation include using different clustering algorithms~\cite{dudoit2003bagging}, using different initializations of the same clustering algorithm or different hyperparameters (e.g. the number of clusters)~\cite{fred2005combining,moderate,kuncheva2006evaluation}, clustering with different subsets of the features~\cite{strehl2002cluster}, using random projections to diversify the feature space~\cite{fern2003random}, and clustering with different subsets of the dataset~\cite{domeniconi2009weighted,dudoit2003bagging}. However, concrete methods for identifying optimal hyperparameters, such as the degree of diversity, the number of clusterings in the ensemble, and the method by which the ensemble is generated, remain elusive.

\textbf{Consensus algorithms:} Consensus algorithms aim to aggregate multiple, diverse clusterings to produce a single, robust solution. Various approaches to this problem have been proposed, such as using matrix factorization~\cite{li2007solving}, distance minimization between clusterings~\cite{zhou2015learning}, utilizing multiple views~\cite{tao2017ensemble}, graph learning~\cite{zhou2021self,huang2017locally,zhou2021tri} and matrix co-association~\cite{jia2022ensemble,huang2018enhanced}. We note that, while improving consensus algorithms increase the robustness of consensus clustering overall, the stages of ensemble generation and  its aggregation with consensus algorithms are largely independent.

\textbf{Consensus Clustering \& Deep Learning:} Despite the established advantages of consensus clustering over single-clustering approaches, consensus clustering has not been explored in the context of deep clustering. A possible reason is the computational cost of generating multiple, diverse base clusterings, which would require training multiple models. The only work that has, to the best of our knowledge, applied consensus clustering in the deep clustering setting is DeepCluE~\cite{huang2022deepclue}. Notably, however, the base clusterings used by DeepCluE are not all learned by the model. Rather, a single-clustering model is trained, and an ensemble is generated by clustering features from multiple layers of the model with U-SPEC~\cite{huang2019ultra}. Our work addresses this gap, by proposing a way to train a single deep clustering model to generate multiple clusterings with controlled diversity and with minimal computational overhead.

\section{Method}\label{sec:Method}

\textbf{Overview:} Our method consists of two components: a) A novel loss function that can be incorporated in deep clustering frameworks to control inter-clustering diversity by applying a threshold to cluster-wise similarities, and b) a method for dynamically estimating that threshold so that the clusterings learned by the model are sufficiently diverse, according to a user-defined metric.

More concretely, we assume a deep clustering model that learns $K$ clusterings (typically a backbone encoder followed by $K$ projection heads), a deep clustering framework and its loss function $L_{main}$, and a diversity target $D^T$ set by the user, expressed as an upper bound to inter-clustering similarity\footnote{It is trivial to modify our formulation to enforce a lower bound. However, experiments (see~\cref{sec:results}) showed that, when learning multiple clusterings, deep clustering frameworks inherently tend to converge to near-identical solutions, which made the lower bound scenario redundant.} (i.e. the maximum acceptable similarity). In order to control the inter-clustering similarity $D^R$ of the learned clusterings so that $D^T\leq D^R$, we propose a complementary loss $L_{div}$. Specifically, given soft cluster assignments for a pair of clusterings $A,B \in K$, we define the inter-clustering similarity matrix $S_{AB}\in \mathbb{R}^{C_A\times C_B}$, where $C_A$ and $C_B$ is the number of clusters in each clustering, and $S_{AB}(i,j)\in [0,1]$ measures the similarity between clusters $i\in C_A$ and $j\in C_B$. It follows that decreasing the values of $S_{AB}$ reduces the similarity between the clusters of $A$ and $B$, and therefore increases their diversity. Accordingly, $L_{div}$ utilizes $S_{AB}$ in order to restrict inter-clustering similarity to be under an upper similarity bound $d$. The value of $d$ is dynamically adjusted during training, decreasing when $D^R>D^T$ and increasing when $D^R\leq D^T$, thereby tightening and relaxing the loss function so that, overall and throughout training, inter-clustering similarity $D^R$ remains at or under the desired level $D^T$.

\begin{figure*}
\centering
\begin{subfigure}[b]{.28\linewidth}
\begin{center}
\includegraphics[width=0.8\linewidth]{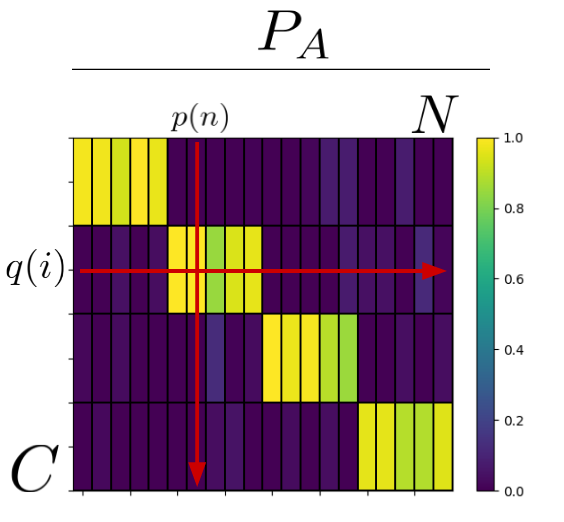}
\end{center}
\caption{Cluster assignments $P_A$ for clustering A}\label{fig:PA}
\end{subfigure}
\begin{subfigure}[b]{.28\linewidth}
\begin{center}
\includegraphics[width=0.8\linewidth]{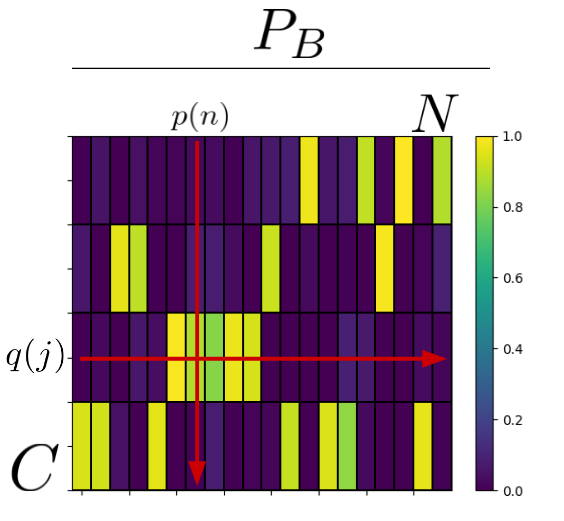}
\end{center}
\caption{Cluster assignments $P_B$ for clustering B}\label{fig:PB}
\end{subfigure}
\begin{subfigure}[b]{.28\linewidth}
\begin{center}
\includegraphics[width=0.8\linewidth]{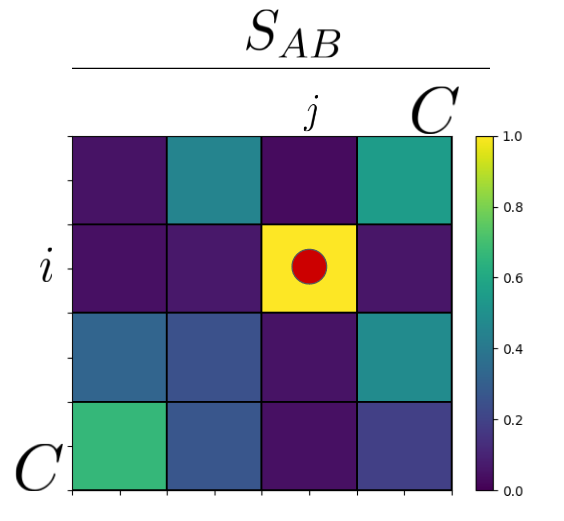}
\end{center}
\caption{Similarity matrix $S_{AB}$}\label{fig:SAB}
\end{subfigure}
\caption{Examples of synthetic cluster assignments $P_A$, $P_B$ and similarity matrix $S_{AB}$. Note that clusters $i\in A$ and $j\in B$ are softly assigned the same samples. Correspondingly, their similarity score $S_{AB}(i,j)$ is high (highlighted with red in~\cref{fig:SAB}). Best seen in color.}
\label{fig:illustrations}
\end{figure*}

\textbf{Defining the inter-clustering similarity matrix:} Our method assumes a standard deep clustering architecture, consisting of an encoder $f$, followed by $K$ projection heads $h_1,..., h_K$, each of which produces assignments for a clustering $k$. Specifically, let $X$ be a set of $N$ unlabeled samples. The encoder maps each sample $x \in X$ to a representation $f(x)$, and each projection head $h_k$ maps $f(x)$ to $C_k$ clusters, so that $p_k(x)=h_k(f(x)) \in \mathbb{R}^{C_k\times 1}$ represents a probability assignment vector mapping sample $x \in X$ to $C_k$ clusters in clustering $k$. Without loss of generality, we assume that $C=C_k\forall k\in K$. Each clustering can then be represented by a cluster assignment matrix $P_k(X)=[p_k(x_1), p_k(x_2), ..., p_k(x_N)]\in \mathbb{R}^{C\times N}$. The column $p_k(n)$, that is the probability assignment vector for the $n$-th sample, encodes the degrees to which sample $x_n$ is assigned to different clusters. The row vector $q_k(i)~\in \mathbb{R}^{N}$ shows which samples are softly assigned to cluster $i\in C$. We refer to $q_k(i)$ as the cluster membership vector.

To quantify the similarity between clusterings $A$ and $B$ we define the inter-clustering similarity matrix $S_{AB} \in \mathbb{R}^{C\times C}$. We define each element $S_{AB}(i,j)$ as the cosine similarity between the cluster membership vector $q_A(i)$ of cluster $i\in A$ and the cluster membership vector $q_B(j)$ of cluster $j\in B$:
\begin{align}
S_{AB}(i,j)=\frac{q_A(i)\cdot q_B(j)}{||q_A(i)||_2||q_B(j)||_2}
\label{eq:sab}
\end{align}
This measure expresses the degree to which samples in the dataset are assigned similarly to clusters $i$ and $j$. Specifically, $S_{AB}(i,j)=0$ if $q_A(i)\perp q_B(j)$ and $S_{AB}(i,j)=1$ if $q_A(i)=q_B(j)$. It is, therefore, a differentiable measure of the similarity of clusters $i$ and $j$.

\textbf{Defining the loss function:} Based on the inter-clustering similarity matrix $S_{AB}$, we define DivClust's loss to softly enforce that a clustering $A$ does not have an \textit{aggregate} cluster similarity with a clustering $B$ greater than a similarity upper bound $d$. The aggregate similarity $S_{AB}^{aggr}$ is defined as the average similarity of clustering $A$'s clusters with their most similar cluster of clustering $B$ (\cref{eq:s_aggr}). Using this metric, we propose $L_{div}$ (\cref{eq:global_cluster}), a loss that regulates diversity between clusterings $A$ and $B$ by forcing that $S_{AB}^{aggr}<d$, for $d\in [0,1]$. It is clear from~\cref{eq:global_cluster} that $S_{AB}^{aggr}<d\Rightarrow L_{div}(A,B)=0$, in which case the diversity requirement is satisfied and the loss has no impact. Conversely, $S_{AB}^{aggr}\geq d\Rightarrow L_{div}(A,B)>0$, in which case the loss requires that inter-clustering similarity decreases.

\begin{equation}
  S_{AB}^{aggr}=\frac{1}{C}\sum_{i=1}^{C}{\underset{j}{max}(S_{AB}(i,j))}
\label{eq:s_aggr}
\end{equation}

\begin{equation}
  L_{div}(A,B)=\left[S_{AB}^{aggr}-d\right]_{+}
\label{eq:global_cluster}
\end{equation}

Having defined the diversity loss $L_{div}$ between two clusterings, we extend it to multiple clusterings $K$ and combine it with the base deep clustering framework's objective. For a clustering $k\in K$, we denote with $L_{main}(k)$ the loss of the base deep clustering framework for that clustering, and with $L_{div}(k,k')$ the diversity controlling loss between clusterings $k$ and $k'$. We present the joint loss $L_{joint}(k)$ for each clustering $k$ in~\cref{eq:total_loss_k}, where $L_{main}(k)$ depends on cluster assignment matrix $P_k$, while $L_{div}(k,k')$ depends on $P_k$ and $P_{k'}$. Accordingly, the model's training loss $L_{total}$, seen in \cref{eq:total_loss}, is the average of $L_{joint}$ over all clusterings.

\begin{equation}
  L_{joint}(k)=L_{main}(k) + \frac{1}{K-1}\sum_{k'=1,k'\neq k}^{K}{L_{div}(k,k')}
\label{eq:total_loss_k}
\end{equation}

\begin{equation}
  L_{total}=\frac{1}{K}\sum_{k=1}^{K}{L_{joint}(k)}
\label{eq:total_loss}
\end{equation}

The loss $L_{total}$ is therefore a combination of the base deep clustering framework's loss $L_{main}$ for each clustering $k\in K$ and the loss $L_{div}$, which is used to control inter-clustering diversity. The proposed loss formulation is applicable to any deep clustering framework that produces cluster assignments through the model (as opposed to frameworks using offline methods such as MIX'EM~\cite{MIXEM}), which covers the majority of deep clustering frameworks outlined in~\cref{sec:related_works}.

\textbf{Dynamic upper bound \textit{d}:} The proposed loss $L_{div}$ controls inter-clustering diversity by restricting the values of $S_{AB}$ according to the similarity upper bound $d$. However, the values of $S_{AB}$ are calculated based on the cosine similarity of \textit{soft} cluster assignments. This means that pairs of cluster assignment vectors $i$, $j$ will have different similarity values $S_{AB}(i,j)$ depending on their sharpness, even if they point to the same cluster in terms of their corresponding hard assignment. It follows that $S_{AB}$ and, accordingly, the impact of $d$, are dependent on the confidence of cluster assignments and vary throughout training and between experiments (as factors like the number of clusters and model capacity influence the confidence of cluster assignments). Therefore, $d$ is an ambiguous and unintuitive metric for users to define diversity targets with.

To tackle this issue and to provide a reliable and intuitive method for defining diversity objectives, we propose dynamically determining the value of the threshold $d$ during training. Concretely, let $D$ be an inter-clustering similarity metric chosen by the user. In this work, we use avg. Normalized Mutual Information (NMI), a well established metric for estimating inter-clustering similarity.

 \begin{equation}
    D = \frac{1}{(K-1)(K/ 2)}\sum_{k=1}^{K-1}{\sum_{k'=k+1}^{K}{NMI(P_k^h,P_{k'}^h)}}
    \label{eq:aggr_nmi}
  \end{equation}
where $P_k^h\in \mathbb{Z}^{N}$ is the hard cluster assignment vector for $N$ samples in clustering $k\in K$ and $NMI(P_k^h,P_{k'}^h)$ represents the NMI between $k$ and $k'$. $D\in [0,1]$, with higher values indicating more similar clusterings.

Assuming a user-defined similarity target $D^T$, expressed as a value of metric $D$, we denote with $D^R$ the measured inter-clustering similarity of the clusterings learned by the model, expressed in the same metric. DivClust's objective is to control inter-clustering diversity, which translates to learning clusterings such that $D^R\leq D^T$. Accordingly, appropriate thresholds $d$ must be used during training. Under the assumption that $D^R$ decreases monotonically w.r.t. $d$, we propose the following update rule for $d$: 
 \begin{equation}
    d_{s+1}=
    \begin{cases}
      max(d_s(1-m),0), & \text{if}\ D^R> D^T \\
      min(d_s(1+m),1), & \text{if}\ D^R\leq D^T
    \end{cases} ,
   \label{eq:calc_d}
   \end{equation}
where $d_s$ and $d_{s+1}$ are the values of the threshold $d$ for the current and the next steps, and $m\in (0,1)$ regulates the magnitude of the update steps. Following this update rule, we decrease $d$ when the measured inter-clustering similarity $D^R$ needs to decrease, and increase it otherwise. For computational efficiency, instead of calculating $D^R$ over the entire dataset in every training step, we do so every 20 iterations on a memory bank of $M=10,000$ cluster assignments -- the latter is updated at every step in a FIFO manner. We set the hyperparameter $m$ to $m=0.01$ in all experiments.

\setlength{\tabcolsep}{10pt}

\begin{table*}[t]
\centering
\begin{small}
\begin{tabular}{c | c c | c c c c c}
\toprule
Framework & Clusterings & $D^T$ & $D^R$ & CNF & Mean Acc. & Max. Acc. & DivClust Acc. \\
\midrule
 \multirow{6}{*}{IIC} & 1 & - & - & 0.997 & 0.442 & 0.442 & 0.442 \\
& 20 & 1. & 0.983 & 0.996 & 0.526 & 0.526 & 0.526 \\
 & 20 & 0.95 & 0.939 & \textbf{0.998} & 0.531 & 0.537 & 0.533 \\
 & 20 & 0.9 & 0.888 & 0.997 & 0.568 & 0.59 & 0.578 \\
 & 20 & 0.8 & 0.8 & 0.997 & \textbf{0.611} & \textbf{0.678} & 0.653 \\
 & 20 & 0.7 & 0.694 & 0.996 & 0.566 & 0.637 & \textbf{0.685} \\
 \midrule
 \multirow{6}{*}{PICA} & 1 & - & - & \textbf{0.906} & 0.533 & 0.533 & 0.533 \\
& 20 & 1. & 0.991 & 0.814 & 0.597 & 0.597 & 0.596 \\
 & 20 & 0.95 & 0.931 & 0.826 & 0.624 & 0.631 & 0.625 \\
 & 20 & 0.9 & 0.891 & 0.841 & \textbf{0.648} & 0.665 & 0.652 \\
 & 20 & 0.8 & 0.817 & 0.828 & 0.598 & 0.635 & 0.595 \\
 & 20 & 0.7 & 0.703 & 0.824 & 0.625 & \textbf{0.691} & \textbf{0.671} \\
 \midrule
 \multirow{6}{*}{CC} & 1 & - & - & \textbf{0.936} & 0.764 & 0.764 & 0.764 \\
& 20 & 1. & 0.976 & 0.934 & 0.763 & 0.763 & 0.763 \\
 & 20 & 0.95 & 0.946 & 0.934 & 0.762 & 0.773 & 0.76 \\
 & 20 & 0.9 & 0.9 & 0.931 & \textbf{0.794} & 0.818 & 0.789 \\
 & 20 & 0.8 & 0.814 & 0.93 & 0.762 & \textbf{0.847} & \textbf{0.819} \\
 & 20 & 0.7 & 0.699 & 0.927 & 0.703 & 0.818 & 0.815 \\
 \bottomrule
\end{tabular}
\end{small}
\caption{Results for IIC, PICA and CC applied on CIFAR10 with DivClust. CNF and Mean Acc. are calculated by averaging the corresponding metrics over all clusterings, while Max Acc. refers to the best performing base clustering. The DivClust Acc. metric measures the accuracy of a consensus clustering produced with the \textit{DivClust C} method.}
\label{tab:diversity_control}
\end{table*}

\setlength{\tabcolsep}{4.5pt}

\begin{table}[t]
    \centering
    \begin{small}
    \begin{tabular}{c | c c c c c}
    \toprule
        \multirow{2}{*}{$D^T$} & \multicolumn{4}{|c}{$D^R$} \\ \cmidrule{2-5}
         & CIFAR10 & CIFAR100 & ImageNet-10 & ImageNet-Dogs  \\ \midrule
        1. & 0.976 & 0.939 & 0.987 & 0.941 \\ 
        0.95 & 0.946 & 0.926 & 0.948 & 0.945 \\ 
        0.9 & 0.9 & 0.848 & 0.897 & 0.87 \\ 
        0.8 & 0.814 & 0.806 & 0.807 & 0.795 \\ 
        0.7 & 0.699 & 0.705 & 0.696 & 0.702 \\ 
        \bottomrule
    \end{tabular}
    \caption{Avg. inter-clustering similarity scores $D^R$  for clustering sets produced by DivClust combined with CC for various diversity targets $D^T$. The objective of DivClust is that $D^R\leq D^T$.}\label{tab:cc_diversity}
    \end{small}
\end{table}

\section{Experiments}\label{sec:experiments}

We conduct several experiments to evaluate DivClust's adaptability, its effectiveness in controlling diversity, and the quality of the resulting clusterings. First, to show that DivClust effectively controls inter-clustering diversity and produces high quality clusterings with various frameworks, we combine it with IIC~\cite{IIC}, PICA~\cite{PICA} and CC~\cite{li2021contrastive}, and apply it to CIFAR10 with various diversity targets $D^T$. Subsequently, we focus on the best framework of the three, namely CC, and conduct experiments on 4 datasets (\mbox{CIFAR10}, CIFAR100, Imagenet-10 and Imagenet-Dogs). Our findings demonstrate that, across frameworks and datasets, DivClust can: a) effectively control diversity and b) improve clustering outcomes over the base frameworks and alternative ensembling methods.

\subsection{Experiments setup}

\setlength{\tabcolsep}{6.5pt}
\begin{table*}[t]
\centering
\begin{small}
\begin{tabular}{c|c|c c c | c c c|c c c|c c c}
\toprule
 Dataset & $D^T$ & \multicolumn{3}{|c|}{CIFAR10} & \multicolumn{3}{|c|}{CIFAR100} & \multicolumn{3}{|c|}{ImageNet-10} & \multicolumn{3}{|c}{ImageNet-Dogs} \\
 \midrule
  Metric & NMI & NMI & ACC & ARI & NMI & ACC & ARI & NMI & ACC & ARI & NMI & ACC & ARI \\
  \midrule
K-means~\cite{kmeans} & - & 0.087 & 0.229 & 0.049 & 0.084 & 0.130 & 0.028 & 0.119 & 0.241 & 0.057 & 0.55 & 0.105 & 0.020 \\
AC~\cite{AC} & - & 0.105 & 0.228 & 0.065 & 0.098 & 0.138 & 0.034 & 0.138 & 0.242 & 0.067 & 0.037 & 0.139 & 0.021 \\
NMF~\cite{NMF} & - & 0.081 & 0.190 & 0.034 & 0.079 & 0.118 & 0.026 & 0.132 & 0.230 & 0.065 & 0.044 & 0.118 & 0.016 \\
AE~\cite{AE} & - & 0.237 & 0.314 & 0.169 & 0.100 & 0.165 & 0.048 & 0.210 & 0.317 & 0.152 & 0.104 & 0.185 & 0.073 \\
DAE~\cite{DAE} & - & 0.251 & 0.297 & 0.163 & 0.111 & 0.151 & 0.046 & 0.206 & 0.304 & 0.138 & 0.104 & 0.190 & 0.078 \\
DCGAN~\cite{DCGAN} & - & 0.265 & 0.315 & 0.176 & 0.120 & 0.151 & 0.045 & 0.225 & 0.346 & 0.157 & 0.121 & 0.174 & 0.078 \\
DeCNN~\cite{DeCNN} & - & 0.240 & 0.282 & 0.174 & 0.092 & 0.133 & 0.038 & 0.186 & 0.313 & 0.142 & 0.098 & 0.175 & 0.073 \\
VAE~\cite{VAE} & - & 0.245 & 0.291 & 0.167 & 0.108 & 0.152 & 0.040 & 0.193 & 0.334 & 0.168 & 0.107 & 0.179 & 0.079 \\
JULE~\cite{JULE} & - & 0.192 & 0.272 & 0.138 & 0.103 & 0.137 & 0.033 & 0.175 & 0.300 & 0.138 & 0.054 & 0.138 & 0.028  \\
DEC~\cite{DEC} & - & 0.257 & 0.301 & 0.161 & 0.136 & 0.185 & 0.050 & 0.282 & 0.381 & 0.203 & 0.122 & 0.195 & 0.079 \\
DAC~\cite{DAC} & - & 0.396 & 0.522 & 0.306 & 0.185 & 0.238 & 0.088 & 0.394 & 0.527 & 0.302 & 0.219 & 0.275 & 0.111 \\
ADC~\cite{ADC} & - & - & 0.325 & - & - & 0.160 & - & - & 0.530 & - & - & - & - \\
DDC~\cite{DDC} & - & 0.424 & 0.524 & 0.329 & - & - & - & 0.433 & 0.577 & 0.345 & - & - & - \\
DCCM~\cite{DCCM} & - & 0.496 & 0.623 & 0.408 & 0.285 & 0.327 & 0.173 & 0.608 & 0.710 & 0.555 & 0.321 & 0.383 & 0.182 \\
IIC~\cite{IIC} & - & - & 0.617 & - & - & 0.257 & - & - & - & - & - & - & - \\
PICA~\cite{PICA} & - & 0.591 & 0.696 & 0.512 & 0.310 & 0.337 & 0.171 & 0.802 & 0.870 & 0.761 & 0.352 & 0.352 & 0.201 \\
CC~\cite{li2021contrastive} & - & 0.705 & 0.790 & 0.637 & 0.431 & 0.429 & 0.266 & 0.859 & 0.893 & 0.822 & 0.445 & 0.429 & 0.274 \\
   \midrule
CC-Kmeans & - & 0.654 & 0.698 & 0.523 & 0.429 & 0.405 & 0.235 & 0.792 & 0.841 & 0.669 & 0.457 & 0.444 & 0.284 \\
   CC-Kmeans/S & - & 0.674 & 0.69 & 0.554 & 0.428 & 0.402 & 0.228 & 0.792 & 0.842 & 0.673 & 0.456 & 0.444 & 0.283 \\
CC-Kmeans/F & - & 0.684 & 0.762 & 0.599 & 0.438 & 0.409 & 0.210 & 0.797 & 0.847 & 0.685 & 0.458 & 0.444 & 0.285 \\
DeepCluE~\cite{huang2022deepclue} & - & \textbf{0.727} & 0.764 & 0.646 & \textbf{0.472} & \textbf{0.457} & \textbf{0.288} & 0.882 & 0.924 & 0.856 & 0.448 & 0.416 & 0.273 \\
   \midrule
\multirow{5}{*}{\textbf{DivClust C}} & 1. & 0.678 & 0.763 & 0.604 & 0.418 & 0.424 & 0.257 & \underline{0.86} & \underline{0.895} & \underline{0.825} & \underline{0.459} & \underline{0.451} & \underline{0.298} \\

 & 0.95 & 0.677 & 0.76 & 0.602 & 0.431 & \underline{0.434} & \underline{0.276} & \textbf{\underline{0.891}} & \textbf{\underline{0.936}} & \textbf{\underline{0.878}} & \underline{0.461} & \underline{0.451} & \underline{0.297} \\

 & 0.9 & 0.678 & \underline{0.789} & \underline{0.641} & 0.422 & 0.426 & 0.258 & \underline{0.879} & \underline{0.92} & \underline{0.859} & \underline{0.48} & \underline{0.487} & \underline{0.332} \\

 & 0.8 & \underline{0.724} & \textbf{\underline{0.819}} & \textbf{\underline{0.681}} & 0.422 & 0.414 & 0.26 & \underline{0.879} & \underline{0.918} & \underline{0.851} & \underline{0.458} & \underline{0.448} & \underline{0.296} \\

 & 0.7 & \underline{0.71} & \underline{0.815} & \underline{0.675} & \underline{0.44} & \underline{0.437} & \underline{0.283} & 0.85 & \underline{0.90} & 0.819 & \textbf{\underline{0.516}} & \textbf{\underline{0.529}} & \textbf{\underline{0.376}} \\
 \bottomrule

\end{tabular}
\end{small}
\caption{Results combining DivClust with CC for various diversity targets $D^T$. We \underline{underline} DivClust results that outperform the single-clustering baseline CC, and note with \textbf{bold} the best results for each metric across all methods and diversity levels. We emphasize that the NMI in this table measures the similarity between the single clustering produced by each method and the ground truth classes. The NMI values representing inter-clustering similarity $D^R$ in ensembles produced by DivClust for the same experiments are presented in~\cref{tab:cc_diversity}.}
\label{tab:detailed_accuracy}
\end{table*}
\setlength{\tabcolsep}{5pt}

\textbf{Datasets:} We conduct experiments with 4 standard datasets in deep clustering: \mbox{CIFAR10}, CIFAR100~\cite{cifar} (evaluating on the 20 superclasses), ImageNet-10 and ImageNet-Dogs~\cite{DAC}.

\textbf{Metrics:} Inter-clustering similarity is measured by averaging the NMI between clusterings to calculate the inter-clustering NMI metric $D$ (\cref{eq:aggr_nmi}), with higher values indicating more similar clusterings. We denote with $D^T$ the diversity target set by the user and with $D^R$ the measured inter-clustering similarity after training. When DivClust is applied we want that $D^R\leq D^T$. Clustering quality is evaluated based on overlap of the clusterings with the dataset's ground truth labels, using the Accuracy (ACC), Normalized Mutual Information (NMI) and Adjusted Rand Index (ARI) metrics. We also report the avg. cluster assignment confidence (CNF), which measures cluster separability. For all four metrics greater values are better, $1$ being optimal.

\textbf{Implementation \& Training:} DivClust is incorporated into the base frameworks as described in~\cref{sec:Method}, by adding DivClust's loss to their objective and duplicating projection heads $h$ to produce multiple clusterings. The models were trained following the configurations (model architecture, training duration, hyperparameters etc.) suggested in their respective papers~\cite{li2021contrastive,PICA,IIC}, unless stated otherwise. PICA and IIC were trained without overclustering. We set the number of clusterings to $K=20$, following convention in consensus clustering~\cite{zhou2021self}, and the number of clusters $C$ to the number of classes for each dataset, following convention for deep clustering evaluation~\cite{li2021contrastive,PICA,IIC}.

\textbf{Consensus Clustering:} To extract single clustering solutions we examine three methods: a) selecting the clustering $k$ with the lowest corresponding loss $L_{main}(k)$ (\textbf{DivClust A}), b) using the consensus clustering algorithm SCCBG~\cite{zhou2021self} to aggregate clusterings (\textbf{DivClust B}), and c) a combination of the two, where we select the 10 best clusterings with regard to their loss and then apply SCCBG (\textbf{DivClust C}). For clarity and space, we present in the paper results only for the hybrid aggregation method \textbf{DivClust C}, which we found to be the most robust. Detailed results for all three approaches are provided in supplementary Tab. 5.

\subsection{Results}\label{sec:results}

Initially, we apply IIC, PICA and CC on CIFAR10, and present the outcomes in~\cref{tab:diversity_control}. We find that, for all three frameworks, DivClust effectively controls diversity, as $D^R$ is consistently close to or lower than $D^T$. Furthermore, results indicate that DivClust is \textit{necessary} to produce diverse clusterings in deep clustering frameworks, as, without it, they tend to converge to near identical solutions (when \mbox{$D^T=1$}, $D^R\rightarrow 1$). Regarding cluster separability, assignment confidence $CNF$ remains high for various diversity targets $D^T$, despite the increased complexity of optimizing both the main deep clustering loss and DivClust's objective. Finally, we observe that, for most diversity targets $D^T$, the mean and max. accuracy, as well as the consensus clustering accuracy produced by the aggregation method \textbf{DivClust C}, increase relative to the single clustering model. Notably, consensus clustering accuracy is higher than the mean clustering accuracy for most cases, which highlights the effectiveness of our approach. We stress that identifying clusterings in the ensemble whose performance matches the mean or max. accuracy is not trivial, which is why consensus clustering is necessary to reach a single clustering solution.

Having established that DivClust is effective across frameworks, we focus on CC and apply it on CIFAR10, CIFAR100, ImageNet-Dogs and Imagenet-10. We compare DivClust with the standard implementation of CC, which is trained to learn a single clustering (\textbf{CC}), as well as with alternative methods of ensemble clustering. Specifically, we apply the typical methods of ensemble generation by extracting the features learned by the single-clustering CC model and running K-means 20 times on the entire dataset (\textbf{CC-Kmeans}), on random subsets of the dataset (\textbf{CC-Kmeans/S}) and on random subsets of the feature space (\textbf{CC-Kmeans/F}), following~\cite{ensemble_survey}. In all three cases, SCCBG is used to aggregate the resulting clusterings. Furthermore, we compare with \textbf{DeepCluE}~\cite{huang2022deepclue}, to the best of our knowledge the only other work that examines consensus clustering in the context of deep clustering, and which is also built on top of CC, allowing for a fair comparison. We note that DeepCluE is not mutually exclusive with DivClust, and could be used jointly with it.

Inter-clustering similarity scores $D^R$ for this set of experiments are presented in~\cref{tab:cc_diversity}, where it is seen that DivClust successfully controls diversity. Results for consensus clustering, the main task for which DivClust is intended, are presented in~\cref{tab:detailed_accuracy} for CC across 4 datasets, where we also include results from other deep clustering frameworks for reference. Detailed results, including aggregation methods DivClust A and DivClust B, as well as mean/max scores for DivClust's clustering ensembles, are provided in supplementary Tab. 5. Results in~\cref{tab:detailed_accuracy} demonstrate that, for most diversity targets $D^T$, DivClust outperforms the single-clustering baseline CC and typical ensemble generation methods, and is competitive with the alternative consensus clustering method DeepCluE. Notably, DivClust is competitive with the baseline across diversity levels. This robustness is very significant, given that identifying what properties (including diversity) lead to optimal outcomes in clustering ensembles is an open problem~\cite{moderate,moderate2}.

To summarize, the results of~\cref{tab:diversity_control,tab:detailed_accuracy,tab:cc_diversity} demonstrate that DivClust: a) effectively controls inter-clustering diversity in deep clustering frameworks in accordance with user-defined objectives, b) does not degrade the quality of the clusterings and in fact produces better solutions than single-clustering models, and c), it can be used with consensus clustering to identify single-clustering solutions superior to those of the corresponding single-clustering frameworks.

\section{Discussion}

\subsection{Diversity Control \& Consensus Clustering Performance} Results presented in~\cref{sec:experiments} demonstrate the effectiveness of DivClust both in controlling inter-clustering diversity and in producing clustering ensembles that lead to consensus clustering outcomes superior to single clustering baselines.

Specifically, \cref{tab:diversity_control,tab:cc_diversity} show that the inter-clustering similarity $D^R$ of ensembles produced by DivClust is consistently lower than the targets $D^T$. In the few cases where $D^R\geq D^T$, it is by very small margins (the greatest deviation was +0.017), which may be attributed to our use of a memory bank to estimate $D^R$ for the update rule of the threshold $d$.
Regarding consensus clustering accuracy, despite the sensitivity of consensus clustering to the properties of the ensembles and, specifically, to different inter-clustering diversity levels, our method proves particularly robust to varying the diversity targets $D^T$, outperforming baselines for most settings. This indicates that DivClust learns clusterings with a good quality-diversity trade-off and can be reliably used for consensus clustering.

Finally, we note that DivClust's ability to explicitly control inter-clustering diversity can facilitate future research on the impact of diversity in clustering ensembles and toward methodologies that determine desirable diversity levels for specific settings~\cite{pividori2016diversity}.

\subsection{Complexity \& Computational Cost}
The complexity of DivClust's objective is $O(nK^2C^2)$, where $n$ is the batch size, $K$ is the number of clusterings, and $C$ is the number of clusters in each clustering. Importantly, the cost of DivClust relates only to the computation of the loss and the additional projection heads, and is therefore \textit{fixed} for fixed $n$, $K$, $C$ values, regardless of model size and data dimensionality, which are generally the computational bottleneck in Deep Learning applications. Therefore, DivClust is scalable to large datasets. Finally, we note that, in practice, the computational overhead introduced by DivClust is minimal. Specifically, for experiments in this work, DivClust learned ensembles with $K=20$ clusterings, with training time increasing between 10\%-50\% relative to the time it took to train the baseline single-clustering models. Contrasted with the alternative of training a single-clustering model 20 times (which would not allow for controlling diversity), DivClust provides an efficient approach for applying consensus clustering with deep clustering frameworks. A detailed analysis on complexity and runtimes is provided in supplementary Sec. C.

\section{Conclusion}
We introduce DivClust, a method that can be incorporated into existing deep clustering frameworks to learn multiple clusterings while controlling inter-clustering diversity. To the best of our knowledge, this is the first method that can explicitly control inter-clustering diversity based on user-defined targets, and that is compatible with deep clustering frameworks that learn features and clusters end-to-end. Our experiments, conducted with multiple datasets and deep clustering frameworks, confirm the effectiveness of DivClust in controlling inter-clustering diversity and its adaptability, in terms of it being compatible with various frameworks without requiring modifications and/or hyperparameter tuning. Furthermore, results demonstrate that DivClust learns high quality clusterings, which, in the context of consensus clustering, lead to improved performance compared to single clustering baselines and alternative ensemble clustering methods.

{\setlength{\parindent}{0.0cm}
\textbf{Acknowledgments:} This work was supported by the EU H2020 AI4Media No. 951911 project.
}

{\small
\bibliographystyle{ieee_fullname}
\bibliography{bib}
}

\vfill
\pagebreak
\appendix

\noindent \textbf{\Large{Supplementary Material}}

\section{Hyperparameters \& Hyperparameter \mbox{Tuning}}\label{sec:hyperparameters}

\paragraph{Diversity target \textbf{$D^T$}:} The diversity target $D^T$, set by the user, is used to indicate how diverse the user wants the clusterings learned by DivClust to be. Specifically, given a similarity metric $D$, $D^T$ represents an upper bound to inter-clustering similarity. That is, for a target $D^T$, the expectation is that the measured inter-clustering similarity $D^R$ of the clusterings learned by the model should be $D^R\leq D^T$. In the paper, we measure inter-clustering similarity $D$ with the avg. NMI between pairs of clusterings, as shown in Eq. 6. Other similarity metrics, however, are also applicable, under the assumption that they decrease monotonically as the dynamic threshold $d$ decreases.

Results presented in paper Tab. 3 demonstrate the effectiveness and robustness of DivClust for various diversity targets, both in terms of successfully controlling diversity and in terms of producing good consensus clustering outcomes. We note, however, that, in the context of ensemble clustering, identifying the optimal degree of inter-clustering diversity is an open problem~\cite{moderate,moderate2} and \textit{beyond} the scope of this work, which proposes a robust method for \textit{controlling} diversity in deep clustering frameworks.

\paragraph{Memory bank size $M$:} As mentioned in Sec. 3 of the paper, in order to update the upper similarity threshold $d$, the inter-clustering similarity score $D^R$ of the learned clusterings must be calculated. This can be highly inefficient for large datasets, as this operation can have very high computational cost. Therefore, to mitigate this problem, we measure inter-clustering similarity over a memory bank, rather than over the entire dataset. Specifically, the memory bank stores cluster assignments for the $M$ samples last seen by the model. The size $M$ of the memory bank should be sufficient for the memory bank to contain a representative subset of the dataset, while taking into account the inherent trade-off with regard to performance. In all our experiments we set the size of the memory bank to $M=10,000$, which we find sufficient, as our largest datasets (CIFAR10 and CIFAR100) have 60,000 samples.

\paragraph{Dynamic upper bound update interval $T$:} The dynamic upper bound $d$ is updated regularly, based on the measured inter-clustering similarity $D^R$, estimated over the memory bank. Specifically, it decreases when $D^R>D^T$ and increases otherwise, as outlined in paper Eq. 7. That calculation and the update of $d$ are executed every $T$ steps, set to $T=20$ in all our experiments. Increasing this value would lead to more frequent updates of $d$ and a corresponding increase in the computational cost of DivClust, as the inter-clustering similarity $D^R$ would be measured more times during training. We found that $T=20$ provides frequent enough updates to achieve the desired diversity target $D^T$ across datasets and deep clustering frameworks, with acceptable computational cost.

\paragraph{Upper bound momentum hyperparameter $m$:} This parameter regulates how big the steps of the upper bound threshold $d$ in either direction are, when the diversity target $D^T$ is/is not satisfied. We note that higher values might lead to instability due to large changes in $d$, however we again found that our initial choice of $m=0.01$ worked well across datasets and frameworks.
\\ \\
The default values for the hyperparameters $M$, $T$ and $m$ were fixed and proved robust across datasets and base clustering frameworks. We note that \textit{no hyperparameter tuning} was found to be necessary when incorporating DivClust to the deep clustering frameworks PICA~\cite{PICA}, IIC~\cite{IIC} and CC~\cite{li2021contrastive}, which highlights DivClust's plug-and-play nature. Indeed, other than duplicating the projection heads of each architecture to produce multiple clusterings, in our experiments we used the same hyperparameters as those reported in the respective papers of the base deep clustering frameworks, including the number of training epochs. More specifically, all three frameworks (IIC~\cite{IIC}, PICA~\cite{PICA} and CC~\cite{li2021contrastive}) use a ResNet-34 architecture. IIC and PICA use Sobel preprocessing on all inputs and a linear projection head, while CC uses a 2-layer MLP projection head. CC resizes all images to $224X224$. IIC and CC train for 1,000 epochs, while PICA trains for 200. More details can be found in the respective papers.

\section{Datasets}\label{sec:datasets}
In this section we provide details for the datasets used in this work. We note that, in all cases, we train and evaluate on both the train and test sets, following convention in deep clustering works. A summary of the datasets is provided in~\cref{tab:datasets}.

\begin{table}[t]
    \centering
    \begin{tabular}{c c c c}
    \toprule
Dataset & Samples & Image size  & Classes \\
\midrule
CIFAR10 & 60,000 & 32X32 & 10 \\
CIFAR100 & 60,000 & 32X32 & 100 (20) \\
ImageNet-Dogs & 19,500 & 96X96 & 15 \\
ImageNet-10 & 13,000 & 96X96 & 10 \\
\bottomrule
    \end{tabular}
    \caption{A summary of the datasets used in the paper. We note that for CIFAR100 we use the 20 superclasses for evaluation.}\label{tab:datasets}
\end{table}

\noindent \textbf{CIFAR10~\cite{cifar}:} An image dataset with 60,000 images, split to 50,000 and 10,000 between the train and test sets. The dataset has 10 classes, and the size of the images is 32X32.

\noindent \textbf{CIFAR100~\cite{cifar}:} An image dataset with 60,000 images, split to 50,000 and 10,000 between the train and test sets. The dataset has 100 classes, organized in 20 superclasses, and the size of the images is 32X32. Following previous works, we evaluate with the 20 superclasses.

\noindent \textbf{ImageNet-Dogs~\cite{DAC}:} A dataset consisting of 19,500 images of dogs organized in 15 classes. Samples were extracted from the ImageNet~\cite{deng2009imagenet} dataset, and their size is 96X96.

\noindent \textbf{Imagenet-10~\cite{DAC}:}  A dataset of 13,000 96X96 images in 10 randomly chosen classes, extracted from the ImageNet~\cite{deng2009imagenet} dataset. We note that we use the same classes as previous works~\cite{DAC,li2021contrastive} for fair comparisons.

\section{Complexity \& Runtime}\label{sec:complexity}

\begin{table}[b]
    \centering
    \begin{tabular}{c c c | c c }
    \toprule
K  & $D^T$   & T   & Time (h) & Time Increase (\%) \\ \midrule
1  & 1.  & -   & 39.1      & 0                  \\
20 & 1.  & -   & 40.5      & 3\%               \\
20 & 0.9 & 20  & 44.6      & 14\%               \\
\bottomrule
    \end{tabular}
    \caption{Runtimes of CC, for 1000 epochs, with CIFAR100 and image size 224X224 during training.}\label{tab:original_time}
\end{table}

\begin{table}[b]
\centering
\begin{tabular}{c c c | c c }
\toprule
K  & $D^T$   & T   & Time (s) & Time Increase (\%) \\ \midrule
1  & 1.  & -   & 141      & 0                  \\ 
20 & 1.  & -   & 161      & 14\%               \\ 
20 & 0.9 & 200 & 166      & 17\%               \\ 
20 & 0.9 & 20  & 209      & 48\%               \\ 
\bottomrule
\end{tabular}
\caption{Runtimes of CC, for 10 epochs, with CIFAR100 and image size 32X32 during training.}
\label{tab:time32}
\end{table}

\begin{table*}[t]
    \centering
    \begin{small}
    \begin{tabular}{c | c c c c }
    \toprule
Method  & CIFAR10 & CIFAR100 & ImageNet-10 & ImageNet-Dogs \\ \midrule
IIC\cite{IIC}  & 21  & -   & - & - \\ 
PICA\cite{PICA} & 6.5  & -   & - & - \\
CC\cite{li2021contrastive} & 44 & 44.5 & 14 & 22 \\ 
\bottomrule
    \end{tabular}
\caption{Runtimes in hours for various models and datasets, for 20 clusterings with DivClust, using the experiment configurations proposed in the respective papers.}\label{tab:bad_runtimes}
    \end{small}
\end{table*}

\begin{figure*}[t]
     \centering
     \begin{subfigure}[b]{0.3\textwidth}
         \centering
         \includegraphics[width=1.\linewidth]{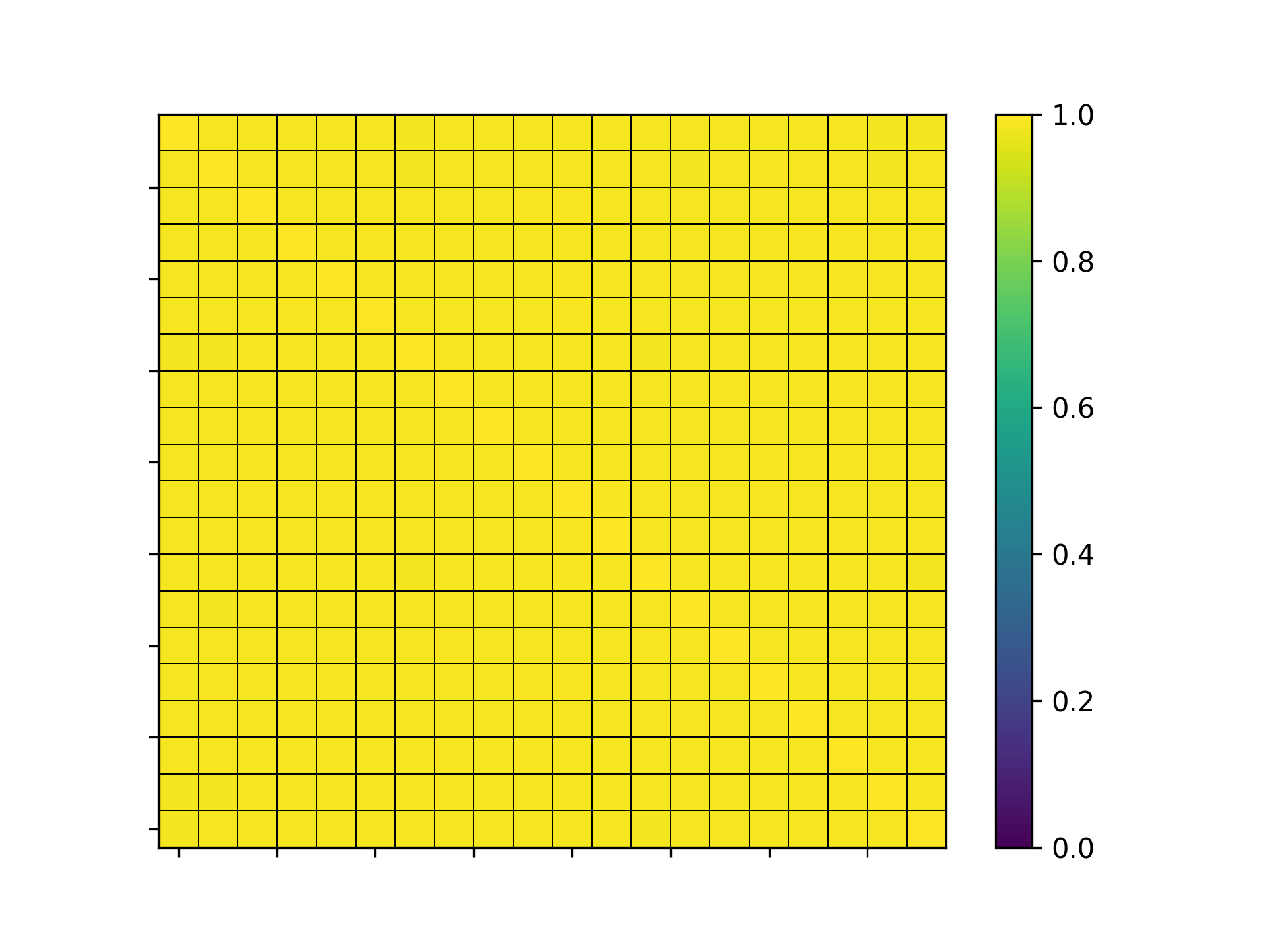}
         \caption{$D^T=1, D^R=0.987$}
         \label{fig:PA1}
     \end{subfigure}
     \begin{subfigure}[b]{0.3\textwidth}
         \centering
         \includegraphics[width=1.\linewidth]{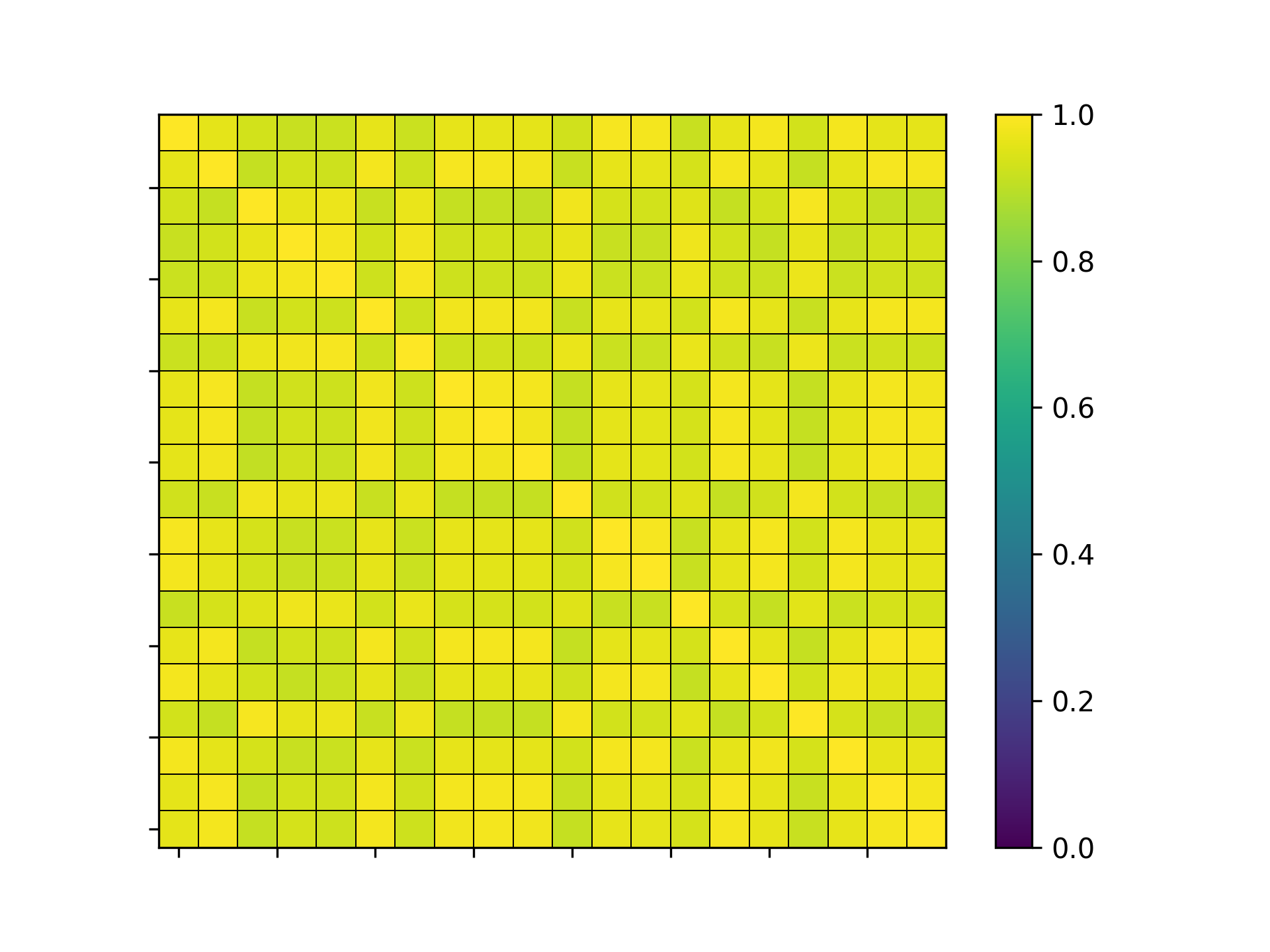}
         \caption{$D^T=0.95, D^R=0.948$}
         \label{fig:PA95}
     \end{subfigure}
     \begin{subfigure}[b]{0.3\textwidth}
         \centering
         \includegraphics[width=1.\linewidth]{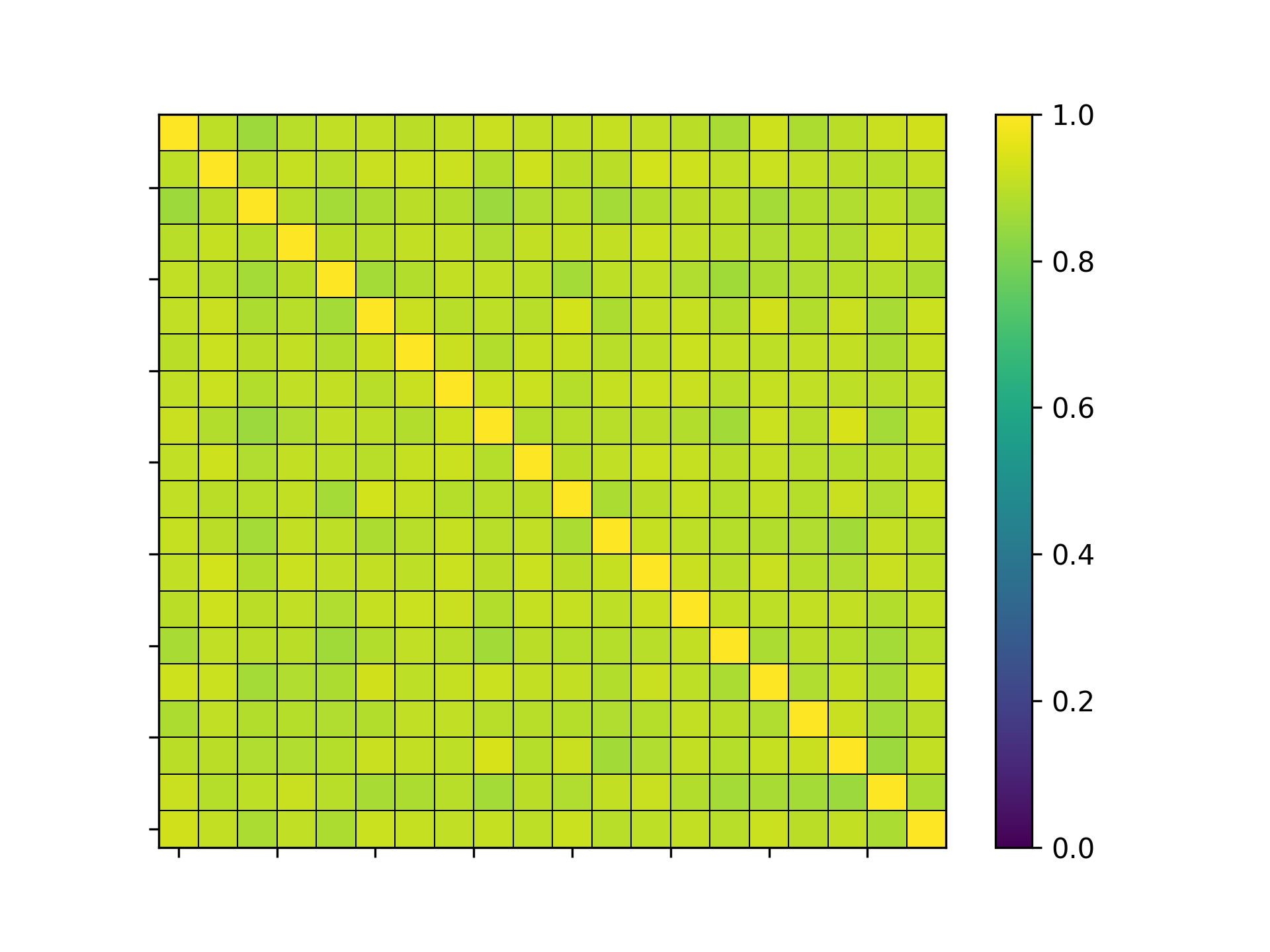}
         \caption{$D^T=0.9, D^R=0.897$}
         \label{fig:PA9}
     \end{subfigure}
     \begin{subfigure}[b]{0.3\textwidth}
         \centering
         \includegraphics[width=1.\linewidth]{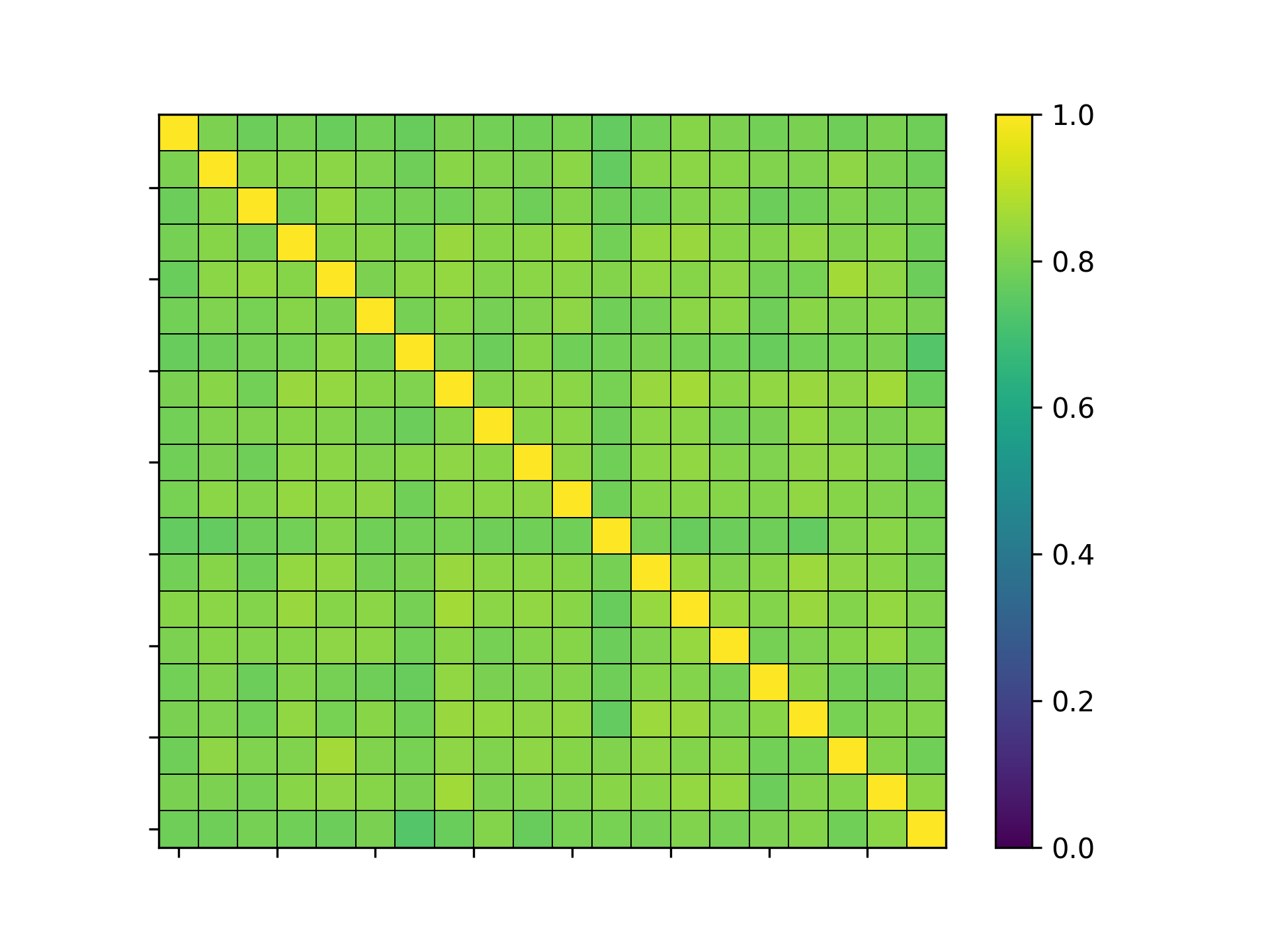}
         \caption{$D^T=0.8, D^R=0.807$}
         \label{fig:PA8}
     \end{subfigure}
     \begin{subfigure}[b]{0.3\textwidth}
         \centering
         \includegraphics[width=1.\linewidth]{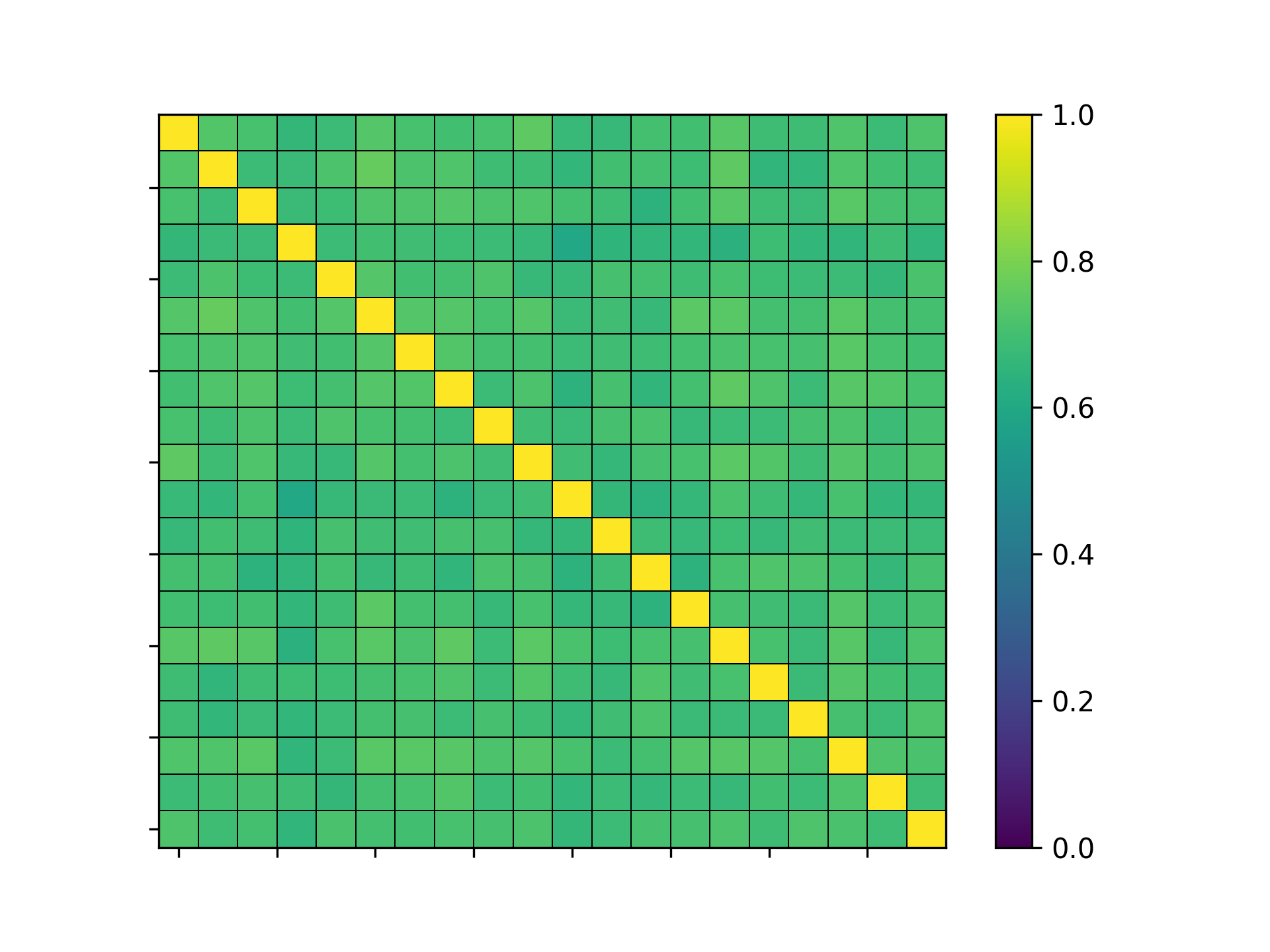}
         \caption{$D^T=0.7, D^R=0.696$}
         \label{fig:PA7}
     \end{subfigure}
        \caption{Visualizations of inter-clustering similarity for ImageNet-10 for various diversity targets $D^T$. Specifically, the heatmaps in each figure represent the NMI between individual clusterings in the corresponding clustering set. For each $D^T$, we also report the measured avg. inter-clustering NMI $D^R$ of the learned clusterings. The figure illustrates how reduced diversity targets $D^T$ (and, accordingly, reduced inter-clustering similarity $D^R$) result in more diverse clusterings. Best seen in color.}
        \label{fig:interclustering_viz}
\end{figure*}

\setlength{\tabcolsep}{5pt}
\begin{table*}[t]
\centering
\begin{small}
\begin{tabular}{c | c | c c c|c c c|c c c|c c c}
\toprule
 Dataset & $D^T$ & \multicolumn{3}{|c|}{CIFAR10} & \multicolumn{3}{|c|}{CIFAR100} & \multicolumn{3}{|c|}{ImageNet-10} & \multicolumn{3}{|c}{ImageNet-Dogs} \\
 \midrule
  Metric & NMI & NMI & ACC & ARI & NMI & ACC & ARI & NMI & ACC & ARI & NMI & ACC & ARI \\
  \midrule
CC-Kmeans & - & 0.654 & 0.698 & 0.523 & 0.429 & 0.405 & 0.235 & 0.792 & 0.841 & 0.669 & 0.457 & 0.444 & 0.284 \\
CC-Kmeans/S & - & 0.674 & 0.69 & 0.554 & 0.428 & 0.402 & 0.228 & 0.792 & 0.842 & 0.673 & 0.456 & 0.444 & 0.283 \\
CC-Kmeans/F & - & 0.684 & 0.762 & 0.599 & 0.438 & 0.409 & 0.210 & 0.797 & 0.847 & 0.685 & 0.458 & 0.444 & 0.285 \\
CC & - & 0.705 & 0.790 & 0.637 & 0.431 & 0.429 & 0.266 & 0.859 & 0.893 & 0.822 & 0.445 & 0.429 & 0.274 \\
DeepCluE & - & 0.727 & 0.764 & 0.646 & 0.472 & 0.457 & 0.288 & 0.882 & 0.924 & 0.856 & 0.448 & 0.416 & 0.273 \\
   \midrule

\textbf{Mean} & \multirow{5}{*}{1.} & 0.678 & 0.763 & 0.604 & 0.418 & 0.427 & 0.257 & 0.859 & \underline{0.895} & \underline{0.824} & \underline{0.457} & \underline{0.451} & \underline{0.297} \\
\textbf{Max} &  & 0.679 & 0.763 & 0.605 & 0.423 & 0.427 & 0.261 & \underline{0.861} & \underline{0.896} & \underline{0.825} & \underline{0.459} & \underline{0.453} & \underline{0.299} \\
\textbf{DivClust A} &  & 0.678 & 0.763 & 0.604 & 0.418 & 0.425 & 0.257 & 0.858 & \underline{0.894} & \underline{0.823} & \underline{0.458} & \underline{0.453} & \underline{0.298} \\
\textbf{DivClust B} &  & 0.678 & 0.763 & 0.604 & 0.418 & 0.424 & \underline{0.267} & 0.858 & \underline{0.895} & \underline{0.823} & \underline{0.459} & \underline{0.452} & \underline{0.298} \\
\textbf{DivClust C} &  & 0.678 & 0.763 & 0.604 & 0.418 & 0.424 & 0.257 & \underline{0.86} & \underline{0.895} & \underline{0.825} & \underline{0.459} & \underline{0.451} & \underline{0.298} \\
   \midrule
   
\textbf{Mean} & \multirow{5}{*}{0.95} & 0.678 & 0.762 & 0.603 & 0.43 & \underline{0.435} & \underline{0.276} & \underline{0.87} & \underline{0.914} & \underline{0.848} & \underline{0.459} & \underline{0.449} & \underline{0.296} \\
\textbf{Max} &  & 0.688 & 0.773 & 0.616 & \underline{0.433} & \underline{0.447} & \underline{0.28} & \underline{0.914} & \underline{0.963} & \underline{0.92} & \underline{0.461} & \underline{0.452} & \underline{0.298} \\
\textbf{DivClust A} &  & 0.683 & 0.768 & 0.61 & 0.43 & \underline{0.434} & \underline{0.276} & \underline{0.916} & \underline{0.964} & \underline{0.922} & \underline{0.452} & \underline{0.461} & \underline{0.298} \\
\textbf{DivClust B} &  & 0.679 & 0.762 & 0.603 & 0.431 & \underline{0.435} & \underline{0.277} & \underline{0.863} & \underline{0.898} & \underline{0.828} & \underline{0.46} & \underline{0.451} & \underline{0.297} \\
\textbf{DivClust C} &  & 0.677 & 0.76 & 0.602 & 0.431 & \underline{0.434} & \underline{0.276} & \underline{0.891} & \underline{0.936} & \underline{0.878} & \underline{0.461} & \underline{0.451} & \underline{0.297} \\
   \midrule

\textbf{Mean} & \multirow{5}{*}{0.9} & 0.703 & \underline{0.794} & \underline{0.644} & 0.422 & \underline{0.43} & 0.262 & \underline{0.861} & \underline{0.903} & \underline{0.832} & \underline{0.471} & \underline{0.479} & \underline{0.323} \\
\textbf{Max} &  & \underline{0.731} & \underline{0.818} & \underline{0.681} & 0.429 & \underline{0.438} & \underline{0.27} & \underline{0.917} & \underline{0.965} & \underline{0.924} & \underline{0.483} & \underline{0.493} &\underline{ 0.34} \\
\textbf{DivClust A} &  & \underline{0.731} & \underline{0.817} & \underline{0.681} & 0.42 & 0.429 & 0.259 & \underline{0.917} & \underline{0.965} & \underline{0.924} & \underline{0.453} & \underline{0.486} & \underline{0.335} \\
\textbf{DivClust B} &  & \underline{0.708} & \underline{0.799} & \underline{0.653} & 0.422 & \underline{0.431} & 0.262 & \underline{0.866} & \underline{0.908} & \underline{0.837} & \underline{0.477} & \underline{0.486} & \underline{0.33} \\
\textbf{DivClust C} &  & 0.678 & 0.789 & \underline{0.641} & 0.422 & 0.426 & 0.258 & \underline{0.879} & \underline{0.92} & \underline{0.859} & \underline{0.48} & \underline{0.487} & \underline{0.332} \\
   \midrule

\textbf{Mean} & \multirow{5}{*}{0.8} & 0.675 & 0.782 & 0.632 & 0.419 & 0.417 & 0.26 & 0.816 & 0.84 & 0.754 & \underline{0.455} & \underline{0.45} & \underline{0.296} \\
\textbf{Max} &  & \underline{0.762} & \underline{0.847} & \underline{0.727} & 0.429 & \underline{0.434} & \underline{0.275} & 0.858 & \underline{0.909} & \underline{0.83} & \underline{0.487} & \underline{0.509} & \underline{0.347} \\
\textbf{DivClust A} &  & \underline{0.762} & \underline{0.847} & \underline{0.727} & 0.419 & 0.42 & \underline{0.275} & 0.835 & 0.845 & 0.779 & \underline{0.486} & \underline{0.504} & \underline{0.347} \\
\textbf{DivClust B} &  & \underline{0.714} & \underline{0.807} & \underline{0.664} & 0.419 & 0.414 & 0.258 & \underline{0.878} & \underline{0.919} & \underline{0.851} & \underline{0.459} & \underline{0.453} & \underline{0.298} \\
\textbf{DivClust C} &  & \underline{0.724} & \underline{0.819} & \underline{0.681} & 0.422 & 0.414 & 0.26 & \underline{0.879} & \underline{0.918} & \underline{0.851} & \underline{0.458} & \underline{0.448} & \underline{0.296} \\
   \midrule

\textbf{Mean} & \multirow{5}{*}{0.7} & 0.645 & 0.703 & 0.556 & 0.43 & 0.425 & \underline{0.267} & 0.742 & 0.747 & 0.643 & \underline{0.458} & \underline{0.453} & \underline{0.298} \\
\textbf{Max} &  & 0.704 & 0.789 & \underline{0.678} & \underline{0.459} & \underline{0.469} & \underline{0.304} & 0.798 & 0.83 & 0.743 & \underline{0.49} & \underline{0.512} & \underline{0.352} \\
\textbf{DivClust A} &  & 0.677 & 0.773 & 0.621 & \underline{0.441} & \underline{0.446} & \underline{0.286} & 0.798 & 0.83 & 0.743 & \underline{0.476} & \underline{0.46} & \underline{0.318} \\
\textbf{DivClust B} &  & 0.665 & 0.725 & 0.621 & \underline{0.434} & \underline{0.438} & \underline{0.272} & \underline{0.875} & \underline{0.916} & \underline{0.837} & \underline{0.492} & \underline{0.456} & \underline{0.315} \\
\textbf{DivClust C} &  & \underline{0.71} & \underline{0.815} & \underline{0.675} & \underline{0.44} & \underline{0.437} & \underline{0.283} & 0.85 & \underline{0.90} & 0.819 & \underline{0.516} & \underline{0.529} & \underline{0.376} \\
   \bottomrule
   
\end{tabular}
\end{small}
\caption{Results combining DivClust with CC for various diversity targets $D^T$ and for various methods of extracting single clustering solutions. We \underline{underline} DivClust results that outperform the single-clustering baseline CC.}
\label{tab:detailed_accuracy_full}
\end{table*}
\setlength{\tabcolsep}{5pt}

\begin{table*}[t]
\centering
\begin{tabular}{c c c | c c c }
\toprule
Method & Clusterings & $D^T$ & Mean Acc.   & Max. Acc.   & Cons. Acc. \\ \midrule
CC & 1 & - & 0.893 & 0.893 & 0.893 \\ 
CC-20x & 20 & - & 0.891 & 0.895 & 0.894 \\ 
\midrule
\multirow{5}{*}{DivClust} & 20 & 1. & 0.895 & 0.896 & 0.895 \\ 
 & 20 & 0.95 & \textbf{0.914} & 0.963 & \textbf{0.936} \\ 
 & 20 & 0.9 & 0.903 & \textbf{0.965} & 0.92 \\ 
& 20 & 0.8 & 0.84 & 0.909 & 0.918 \\ 
 & 20 & 0.7 & 0.747 & 0.83 & 0.9 \\ 
\bottomrule
\end{tabular}
\caption{Results on Imagenet-10 for the baseline single-clustering method CC, for 20 clusterings learned by training CC 20 times with different seeds (\textbf{CC-20x}), and for DivClust with various diversity targets $D^T$. We note the best results with \textbf{bold}.}
\label{tab:20x}
\end{table*}

\paragraph{Complexity:} As stated in paper Sec. 5, the complexity of DivClust is $O(nK^2C^2)$, where $n$ is the batch size, $K$ is the number of clusterings, and $C$ is the number of clusters in each clustering. Importantly, given fixed hyperparameters $n$, $K$ and $C$, the computational cost of DivClust is fixed, regardless of the size of the model and the dimensionality of the input data. Therefore, DivClust is scalable to large datasets and deep learning architectures.

\paragraph{Runtime Analysis:} To analyze the practical impact of DivClust we first present runtimes with CC\cite{li2021contrastive} on CIFAR100 in~\cref{tab:original_time}. The experiments were conducted with CC's default settings of 1000 epochs and images resized to 224X224 during training. We present results for $K=1$ clustering (the default CC framework), $K=20$ \textit{without} DivClust (where $D^T=1$ so the diversity loss is not used and $d$ is not updated), and $K=20$ \textit{with} DivClust ($D^T=0.9$). The update interval for $d$ is set to the default $T=20$. We note that, in terms of runtime, the specific value of the diversity target $D^T$ does not have an impact, as long as $D^T<1$. To provide a more robust analysis of DivClust's components with regard to their computational cost, in~\cref{tab:time32} we explore the impact of a) the dimensionality of the input data, and b) the frequency of the updates of $d$. Specifically, we train CC for 10 epochs (2,340 steps) with the standard image size for CIFAR100, namely 32X32, and include results for a less frequent update of $d$, where $T=200$. All experiments were conducted on a single RTX6000 GPU.

For completeness, in addition to the experiments of~\cref{tab:original_time,tab:time32}, which were conducted specifically for runtime analysis while ensuring that interference in their machine was kept at a minimum, we present approximate runtime figures for each dataset and framework \textit{with DivClust} in~\cref{tab:bad_runtimes}.

\paragraph{Conclusions: } Based on the complexity of the framework and the results presented in~\cref{tab:original_time,tab:time32}, we note the following: \begin{itemize}
    \item The practical impact of DivClust in terms of increased training time is very small. Specifically, as seen in~\cref{tab:original_time}, CC with DivClust requires 44.6 hours to train, as apposed to 39.1 hours without DivClust (a 14\% increase). For comparison, the alternative of running the model 20 times would require \textit{~32 days}, and would offer no control over the outcome in terms of inter-clustering diversity.
    
    \item Given that the computational cost of DivClust is independent of the model's backbone, its relative impact decreases for larger models and/or input dimensionality, given fixed $n$, $C$ and $K$. That is evident by comparing~\cref{tab:original_time,tab:time32}, where increasing the size of the input images from 32X32 (\cref{tab:time32}) to 224X224 (\cref{tab:original_time}) decreases the relative runtime increase from 48\% to 14\%, as the backbone's load increases while DivClust's remains fixed. This makes DivClust well suited for deep model architectures.

    \item Experiments for $K=20$ \textit{without} DivClust ($D^T=1$) were faster than experiments \textit{with} DivClust ($D^T<1$) by a small margin, which is to be expected. However, as was shown in Sec. 4 of the paper, without DivClust clusterings tend to converge to the same solution. Therefore, this approach is unsuitable for producing multiple, diverse clusterings, and, by extension, unsuitable for consensus clustering.
\end{itemize}

Overall, the computational cost produced by DivClust is very small relative to that of the base deep clustering models. Furthermore, the relative impact of DivClust decreases for larger architectures. Therefore, DivClust can be considered to be a highly efficient and scalable method for producing diverse clusterings in the context of deep clustering. 

\begin{figure*}[t]
    \centering
    \includegraphics[width=1.\linewidth]{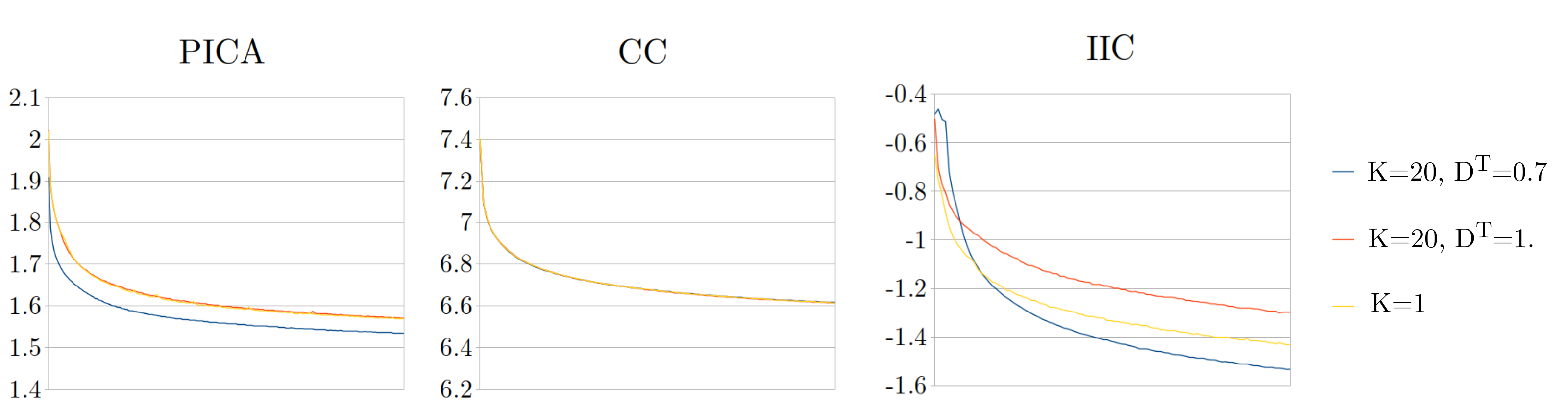}
    \caption{The training loss $L_{total}$ for PICA, CC and ICC, trained on CIFAR10 to learn a single clustering (K=1), multiple clusterings \textit{without} diversity (K=20, $D^T=1$) and multiple clusterings \textit{with} diversity (K=20, $D^T=0.7$). Best seen in color.}
    \label{fig:losses}
\end{figure*}

\section{Visualizing inter-clustering diversity}

To illustrate the impact of DivClust, we present in~\cref{fig:interclustering_viz} visualizations of the diversity between clusterings, for sets of clusterings produced by DivClust. Each subfigure in~\cref{fig:interclustering_viz} corresponds to a set of 20 clusterings produced by DivClust combined with CC, trained on ImageNet-10 for various diversity targets $D^T$. Specifically, the subfigures consist of 20X20 matrices, where each value $(i,j)$ represents the NMI between clusterings $i$ and $j$, with higher values corresponding to more similar clusterings.

In~\cref{fig:interclustering_viz}, one can see that decreasing the diversity target $D^T$ indeed results to less similar clusterings. Furthermore, one can see that the similarities between pairs of clusterings are not uniform. That is, they are not all equally diverse with each other. This reflects the fact that DivClust controls the \textit{avg.} inter-clustering similarity, therefore individual pairs of clusterings may have a higher similarity score than $D^T$, as long as the avg. similarity score $D^R$ is lower than $D^T$. We note that it is trivial to modify DivClust's loss to enforce diversity between each pair of clusterings. However, for the purposes of consensus clustering, the more relaxed constraint of controlling diversity on the aggregate was preferred.

\section{Extended CC results}

In this section, detailed results are presented for experiments combining DivClust with CC. Following the methodology outlined in Section 4 of the paper,~\cref{tab:detailed_accuracy_full} includes results for CIFAR10, CIFAR100, ImageNet-Dogs and ImageNet-10, reported for each of the three proposed methods for extracting single clustering solutions, namely \textbf{DivClust A} (selecting the clustering $k$ with the lowest loss $L_{main}(k)$), \textbf{DivClust B} (applying consensus clustering), and the method we found to be the most robust, \textbf{DivClust C} (selecting the 10 best clusterings in terms of their loss, and applying consensus clustering on them). In~\cref{tab:detailed_accuracy_full}, we also include the mean/max values of each metric over the clustering ensembles produced for each setting, noting that, in practice, identifying clusterings whose performance matches those values is non-trivial, as we assume that we do not have access to the labels.

Finally, in~\cref{tab:20x}, we present results on Imagenet-10 for DivClust trained with various diversity targets $D^T$, comparing it with the single-clustering baseline CC and with a clustering ensemble produced by training a single-clustering model 20 times with different seeds (\textbf{CC-20x}). In all cases, the consensus clustering solution was produced by identifying the 10 best performing clusterings of each set with regard to their loss, and applying the SCCBG~\cite{zhou2021self} consensus clustering algorithm.~\cref{tab:20x} demonstrates that, despite requiring approximately 20X more training time, producing the ensemble from multiple individually trained models leads to minimal performance gains over the baseline, as opposed to DivClust, which consistently outperforms the baseline in terms of consensus clustering accuracy.

\section{Joint optimization and convergence analysis}

To further demonstrate that DivClust can be straightforwardly integrated in deep clustering frameworks, we analyze its behavior with regard to the training loss and its convergence. Specifically, in~\cref{fig:losses}, we present the total loss $L_{main}$ during training for the three deep clustering frameworks CC~\cite{li2021contrastive}, PICA~\cite{PICA} and IIC~\cite{IIC}. The frameworks are applied on CIFAR10 and trained to learn a) a single clustering, b) multiple clusterings (K=20) without diversity requirements ($D^T=1$), and c) multiple clusterings with diversity ($D^T=0.7$).

We observe that different frameworks do not behave in exactly the same way. Specifically, while CC's loss curve remains virtually identical in all three examined cases, PICA and IIC converge to different loss values when DivClust is active (i.e. when $D^T=0.7$). We attribute this to the frameworks' different objectives and architectures. However, in all cases, the loss converges smoothly, which indicates that our proposed loss $L_{div}$ can be optimized jointly with each framework's base loss $L_{main}$ without requiring adjustments and without disturbing the training process.

\end{document}